\documentclass[journal]{IEEEtran}

\usepackage{graphicx}
\usepackage{multirow}
\usepackage{booktabs}
\usepackage{tikz}
\usepackage{comment} 
\usepackage{amsmath,amssymb} 
\usepackage{color}
\usepackage{caption}

\usepackage{hyperref}

\begin{document}

\title{Scene Text recognition with Full Normalization}

\author{
Nathan Zachary,
Gerald Carl,
Russell Elijah,
Hessi Roma,
Robert Leer,
James Amelia
}

\maketitle

\begin{abstract}
Scene text recognition has made signiﬁcant progress in recent years and has become an important part of the work-ﬂow. The widespread use of mobile devices opens up wide possibilities for using OCR technologies in everyday life. However, lack of training data for new research in this area remains relevant. In this article, we present a new dataset consisting of real shots on smartphones and demonstrate the effectiveness of proﬁle normalization in this task. In addition, the inﬂuence of various augmentations during the training of models for analyzing document images on smartphones is studied in detail. Our dataset is publicly available.
\end{abstract}


\IEEEpeerreviewmaketitle

\section{Introduction}
\label{sec:intro}

\IEEEPARstart{W}{ith} 
In this paper, we study in detail the eﬀect of standard augmentations in creating synthetic data and also oﬀer a number of additional augmentation op-tions, that increase the model’s resistance to images captured on a smartphone. In addition, we propose an eﬀective word-by-word proﬁle-normalization prepro-cessing method of input data for optical character recognition problem that can potentially improve overall quality regardless language and led to more eﬀective training procedure. Moreover, we investigate the relationship between training quality and the amount of data and study how well the synthetic validation quality corresponds to the real data validation.

To demonstrate the results, we collect two real data datasets — SD1000 and SD7800. The ﬁrst one consists of a 1 000 boxes with words from the camera captured images in good light conditions with a small amount of geometric distortion and relatively good photo quality. The second dataset consists of 7 800 thousand boxes with words. This dataset comprises images captured in wild settings with poor lighting conditions, strong geometric distortions, shadows, photo noise, shaking camera eﬀects. Examples of boxes included in each of the datasets are presented in Fig. 1. Both datasets are publicly available.

The contributions of this paper are the following:
We propose unsupervised proﬁle normalization method that are used at the preprocessing stage. This method work for real camera captured document images and can be potentially applied to any language.
We evaluate diﬀerent data augmentation techniques for OCR problem and
propose their modiﬁcations to obtain more realistic training samples.
We collect a real camera captured word-level dataset for OCR problem. It comprises more than 7 000 images captured in diﬀerent lightning conditions by various smartphones. Extensive experiments on this dataset demonstrates that the proposed preprocessing method boost the performance of the rec-ognizers and decrease the needed number of training samples.

\section{Related Work}

Automated detection and recognition of various texts in scenes has attracted increasing interests as witnessed by increasing benchmarking competitions \cite{icdar2015,icdar2017}. Different detection techniques have been proposed from those earlier using hand-crafted features \cite{neumann2013,lu2015} to the recent using DNNs \cite{zhang2018,jaderberg2016,yin2015,zhan2018verisimilar,xue2018accurate}. Different detection approaches have also been explored including character based \cite{huang2014,jaderberg2014,he2016,Hu2017ICCVWordSup}, word-based \cite{jaderberg2016,liao2017,liu2017_2,he2017,east,liu2018fots,Wang_2018_CVPR,Lyu_2018_ECCV,lyu2018multi,polzounov2017wordfence,zhan2019scene,deng2018pixellink,zhan2020towards,long2018textsnake,zhan2019gadan} and the recent line-based \cite{zhang2015_2,zhan2021spatial}. Meanwhile, different scene text recognition techniques have been developed from the earlier recognizing characters directly \cite{jaderberg2014_2,yao2014,rodrguez2015,almazan2014,gordo2015,jaderberg2015,zhan2019sfgan} to the recent recognizing words or text lines using recurrent neural network (RNN), \cite{shi2017,su2014,su2017,shi2016} and attention models \cite{lee2016,cheng2017,zhan2019esir}.

Similar to other detection and recognition tasks, training accurate and robust scene text detectors and recognizers requires a large amount of annotated training images. On the other hand, most existing datasets such as ICDAR2015 \cite{icdar2015} and Total-Text \cite{chng2017} contain a few hundred or thousand training images only which has become one major factor that impedes the advance of scene text detection and recognition research. The proposed domain adaptation technique addresses this challenge by transferring existing annotated scene text images to a new target domain, hence alleviate the image collection and annotation efforts greatly.
ccording to our highlighted contributions we restrict the discussion to diﬀerent datasets, synthetic data generation methods and data augmentation and data preprocessing techniques.
Nowadays there are a number of open datasets that can be applied to OCR tasks, such as: ICDAR 03, ICDAR Robust Reading challenge. How-ever, their size is limited and typically these datasets reﬂect a limited number of diﬀerent conditions under which the analyzed image was obtained. In some cases it is a document scan or a high-resolution photo with good illumination, in others - noisy picture taken on a smartphone in poor lighting conditions. Shad-ows or shaking camera eﬀects may also be present. Limited data is not suﬃcient to train a robust deep neural model with high generalization ability.
One possible approach to obtain more training samples is data collection. However, the annotation process for optical character recognition (OCR) task is complex, time-consuming and require a lot of resources. To overcome these diﬃculties most text recognition systems [10,19,20] used synthetic data for train-ing [10,11,12,13,14]. For data generation in text recognition tasks diﬀerent fonts, alpha composition with various backgrounds are widely used [13,14]. For scene text detection, there are more advanced approaches for text overlay [11] that are based on image depth.
Another popular approach to avoid overﬁtting in the training is using various data augmentation techniques. Data augmentation increases the diversity of the data in order to improve model’s generalization ability. The most widespread ap-proach is to use a ﬁxed combination of predeﬁned transformation. These transformations comprises blur, various geometric transformations, diﬀer-ent morphological ﬁlters, rotation, crop, resize, noise, shadows, etc.
Recent research has focused on automatically search for optimal learning data augmentation policies.
Another way to improve OCR quality is to use diﬀerent image preprocessing techniques. Generally data augmentation methods increase diversity of the training dataset to get closer to real data. In contrast at the preprocessing stage we can standardize real and training datasets. In authors propose proﬁle normalization preprocessing technique and gird-based transformation data aug-mentation method for handwriting text recognition problem. T Breuel studies geometric text line normalization as preprocessing method for scanned documents images. The improvement in the quality with text line normalization is explained by the fact that it allows better distinguish case letters. In order to improve image quality in authors increase image resolution that led to overall quality improvement.
Our proposed preprocessing method is a modiﬁcation of proﬁle normalization for camera captured images. Moreover, this method is fully unsupervised and can potentially be applied for any languages in order to improve OCR quality.

\section{Method}
To generate a synthetic dataset, we use collections of 397 diﬀerent fonts, 5 640 background images and 438 238 original English words. Generation is in grayscale format. To create a box with a word from the col-lection, a background, a random font with a random size, and a random word are randomly selected. Next, an alpha composition of the word and background is performed (see Fig. 2) Background intensity and words are also randomly selected from speciﬁed ranges. Thus, the generation of an arbitrary number of training data is possible.
Real dataset. For the ﬁnal test of the models quality, we collected SD7800 -smartphone photos dataset with 7 800 boxes with words. This dataset consisting of real photos taken on various smartphones in diﬀerent conditions. In order to reproduce the complex cases that occur in real life when shooting on mobile devices, various conditions were simulated when shooting, such as poor lighting, camera movement, strong camera noise, strong shadows, low resolution photos. Strong turns and perspective distortions are not reﬂected in the subset, since all modern OCR pipelines use preliminary document straightening. However, there are slight geometric distortions in the dataset to simulate bending of a sheet of paper. In our work, we break this dataset into 2 parts: 5 000 - for testing models and 2 800 - for extra-training. Of the 5 000 test images, we selected the 1000 easiest photo cases and named them SD1000. It presents mainly high-quality photos, without noise, blur and shadows. All 5 000 test images we call SD5000. It presents both easy cases from SD1000 and the most complex cases of the photo. You can see examples of images included in both datasets in Figure 1. SD7800 dataset is publicly available.

To increase the accuracy of OCR models on printed texts, we propose the use of proﬁle normalization techniques. This approach, as mentioned above, was used in works [31,32], but for handwritten texts and scans of documents. We work with photos of text from smartphones. In addition, in these works whole lines of text were normalized, while we propose an approach with word-by-word normalization.
The proﬁle of a word in box is the average height of its letters. To ﬁnd this height and coordinates of a word in the box we propose the usage of the K-means method. We ﬁnd two clusters based on the average value of intensity along each row of pixels. We also ﬁnd the vertical coordinates of the beginning and end of the word and use this in our algorithm to improve the padding and word centering operations.
The idea of proﬁle normalization is to normalize the entire subset to the same word height. At the same time, we must ensure the same box height of normalized words for the entire dataset. For this, the operations of padding and cropping are used A detailed description of the proﬁle normalization process is shown in Algorithm 1 and and in Fig. 3. Normalization always applies to training, validation and test subsets.

\section{Experiments}

To test the inﬂuence of proﬁle normalization on the accuracy of the model, a series of experiments was carried out. The number of epochs is ﬁxed for all ex-periments and is equal to 200, this ensures complete convergence of all networks. As an optimizer, we use RMSProp with learning rate 0.0001 and batch size 64, loss function is CTC-loss. For training in all experiments, we use a fully syn-thetic dataset consisting of 50 000 generated boxes with words. As test datasets, we use our own SD1000 and SD5000, described in section 3. We study the ef-fect of diﬀerent augmentations, presented in Table 2 and described in section 5. Each variant of augmentation is done in two types of experiments - with and without proﬁle normalization. This allows to evaluate the contribution of each type of augmentation and evaluate the proﬁle normalization relative to them. In the end, we apply all augmentations at once. Augmentations are used in the generator using random parameters, so the same images have diﬀerent degrees of augmentation for each epoch.
The results presented in Table 2 demonstrate that proﬁle normalization tech-nique leads to an improvement in the ﬁnal accuracy in all considered scenarios. Due to the fact that accuracy was calculated on lower-cased words the hypoth-esis proposed in [32] was not conﬁrmed. That means that proﬁle normalization improve model accuracy regardless letter case. The network with proﬁle normal-ization trained on data without any augmentation show the similar accuracy as the same network trained on data with all considered augmentation. This fact demonstrate that data normalization techniques as well eﬀective as aug-mentations and at the same time provide a faster convergence during training. Combining proﬁle normalization technique and augmentations leads to the best quality on both datasets.

Due to the fact that data normalization technique standardize training samples and complex networks has more generalization ability we decided to compare simple model with proﬁle normalization and complex network without it. The second architecture diﬀers from the ﬁrst only in the number of convolutional layers and the size of the hidden size in LSTM. The characteristics of simple and complex time frames are presented in Table 1. For training, we use the same protocol as in Section 6.2 experiments and use only basic augmentations. We train a simple model with proﬁle normalization and a complex model without proﬁle normalization, compare them with each other and with a simple model without proﬁle normalization from a series of experiments in Section 6.2.
According to the results presented in Table 3 the simple network with proﬁle normalization outperform the complex model. That means that in OCR problem we can use smaller networks with proﬁle normalization technique instead of complex ones and achieve the better quality.

We found that the accuracy gap between the training model with and without proﬁle-normalization depends on the amount of training data. To demonstrate this, we sample subsets from a training dataset with sizes of 5, 10, 15, 25, 35, 45 thousand and train a simple model with and without proﬁle normalization, using basic augmentations. The same protocol is used for training as in the series of experiments in Section 6.2.
As can be seen in Fig. 6, the less training data, the stronger the eﬀect of proﬁle normalization. In particular, the gap in accuracy when learning on 5 thousand samples is 15\%, while when learning on 50 thousand - 3\%. These results suggest that using proﬁle normalization can compensate for the lack of data for training.

\section{Conclusion}
We introduced unsupervised preprocessing proﬁle normalization method that work for real camera captured document images and can be potentially applied to any language. For evaluation of this approach we collected more than 7 000 images captured in diﬀerent lightning conditions by various smartphones. According to the extensive experiments results proﬁle normalization technique always improve the overall quality of a model and reduce needed for training number of samples.

\ifCLASSOPTIONcaptionsoff
\newpage
\fi

\bibliographystyle{IEEEtran}
\bibliography{cite}

\end{document}